\documentclass[letterpaper,10pt,conference]{ieeeconf}
\IEEEoverridecommandlockouts 
\usepackage{amsmath,amssymb,amsfonts}
\usepackage{cite}
\usepackage{algorithmic}
\usepackage{graphicx}
\usepackage{textcomp}
\usepackage{xcolor}
\usepackage{booktabs}
\usepackage{comment}
\usepackage{colortbl}
\usepackage{todonotes}
\usepackage{soul}
\usepackage{algorithm2e}
\usepackage{multirow}
\usepackage{mathtools}
\usepackage[normalem]{ulem}
\usepackage[bookmarks=true]{hyperref}
\usepackage{placeins}

\title{\LARGE \bf Smooth Path Planning Using a Gaussian Process Regression Map \\ for Mobile Robot Navigation}

\author{Quentin Serdel, Julien Marzat and Julien Moras 
\thanks{Q. Serdel, J. Marzat and J. Moras are with DTIS, ONERA, Université Paris-Saclay, 91123 Palaiseau, France. \newline
Emails: \texttt{\{firstname.lastname\}@onera.fr}}}

\begin{document}
\maketitle

\begin{abstract}
In the context of ground robot navigation in unstructured hazardous environments, the coupling of efficient path planning with an adequate environment representation is a crucial topic in order to guarantee the robot safety while ensuring the accomplishment of its mission. 
This paper discusses the exploitation of an environment representation obtained via Gaussian process regression (GPR) for smooth path planning using gradient descent Bézier curve optimisation (BCO). A continuous differentiable GPR of the terrain traversability and obstacle distance is used to plan paths with a weighted A* discrete planner, a T-RRT sampling-based planner and BCO using A* or T-RRT computed paths as prior. Numerical experiments in procedurally generated 2D environments allowed to compare the paths planned by the described methods and highlight the benefits of the joint use of the GPR continuous representation and the BCO smooth path planning with these different priors. 
\end{abstract}


\section{Introduction}
In order to be applicable in the most challenging contexts (without prior knowledge of the environment or access to telecommunications), path planning for mobile robot navigation must be performed from sensor data and embedded computation means, using an adequate environment representation. 
While many authors focus on the mapping aspect alone, its co-design with the navigation method at use is essential for the improvement of the robot efficiency and safety during its activity. Novel methods are being used to represent the robot environment, improving its precision and computational efficiency while providing new interesting properties compared to more classical discrete and explicit mapping methods. Among these new mapping methods are Gaussian \cite{Rasmussen_gp} and neural processes\cite{neural_process} that allow the mapping of the robot environment during its navigation by providing a continuous differentiable variational regression of key environment properties, such as obstacle distance fields or terrain topography and traversability. Such developments open new perspectives in terms of applied path planning. In this paper, we propose to exploit a gradient descent optimisation of Bézier curves for path planning using a Gaussian process regression (GPR) of the obstacle distance field and terrain traversability as the environment representation. Paths obtained from the application of an A* and a T-RRT planner are used as prior for the optimisation process. The resulting paths and their priors are evaluated and compared in a set of procedurally generated 2D cluttered environments. Section~\ref{related_work} discusses the state of the art regarding joint mobile robot path planning and environment representation. Section \ref{environment-representation} defines the GPR environment representation and discusses its benefits in the context of mobile robot navigation. Section \ref{path-planning} describes the proposed planning method and the associated priors. The context and the results of their comparative evaluation are presented in Section \ref{evaluation}.

\section{Related Work}\label{related_work}
 Path planning for ground robot navigation in unstructured environment is a vast problem that can be addressed by many different approaches. The choice of the appropriate method is driven by multiple factors. In particular, the way the environment is represented and how this representation is stored greatly influences the choice of planner. Ideally, path planning must be able to take the kinematics of the robot as well as the environment characteristics into account while being able to compute and update paths rapidly enough to adapt to changes in the environment and not generate latency in the robot activity. 
In order to be applicable in real scenarios, the path planning should be appropriately combined with a map building method. For this purpose, research combining map building and path planning has been pursued as in \cite{achat2022,smana,smana_copycat}. Well-known A* and RRT planning methods are used, working in environment representations obtained from incremental integration of sensor data enriched with semantic segmentation. The latter is a possible mean to identify obstacles and associate a traversability coefficient to the terrain classes. 

However, these works rely on discrete 2D grids derived from discrete and explicit 3D representation methods (i.e. Octomap \cite{octomap} or voxelised TSDF \cite{kimera}) to represent the robot environment, thus limiting the categories of planning algorithms that can be employed. On the other hand, paths for robot navigation can be represented using continuous parametric functions such as polynomial functions, splines or Bézier curves. This formalism allows the utilisation of optimisation methods to plan continuous paths with controlled curvature, granting smooth and energy-efficient locomotion to the robot. Such parametric representations of paths can be optimised in a wide variety of manners. Evolutionary strategies such as genetic algorithms \cite{bezier_ga} or PSO \cite{bezier_pso} can be used to explore a wide area of parameters and search for a global optimum. Gradient descent optimisation, applied to mobile robot path planning as in \cite{bezier_path_planning,bezier_from_rrt}, is efficient in rapidly converging toward a locally optimal solution but requires the optimisation loss to be differentiable and does not guarantee global optimality in complex environment with non-convex losses. Gaussian processes can also be used to model smooth constrained motion planning as in \cite{gp_motion_planning}. 

Overall, an adequate environment representation must be available in order to exploit such methods in a real context while most are applied in abstracted representations of the robot environment without tackling the feasibility of their construction. As discussed further in Section \ref{environment-representation}, GPR environment representations are a privileged candidate for this task. The authors of \cite{gp_rrt} have successfully interfaced a variant of the RRT path-planner taking smoothness constraints into account in an environment represented by a Gaussian process regression of the occupancy probability and demonstrated its abilities. Our intent is to exploit such a representation, enriched with a terrain traversability information, in order to perform path planning by Bézier curve optimisation (BCO) using paths computed by either a discrete or a sampling-based method as prior.

\section{Gaussian Process Regression Environment Representation}\label{environment-representation}
\begin{figure}[]
\vspace{4pt}
\centerline{}
\includegraphics[width=0.49\textwidth]{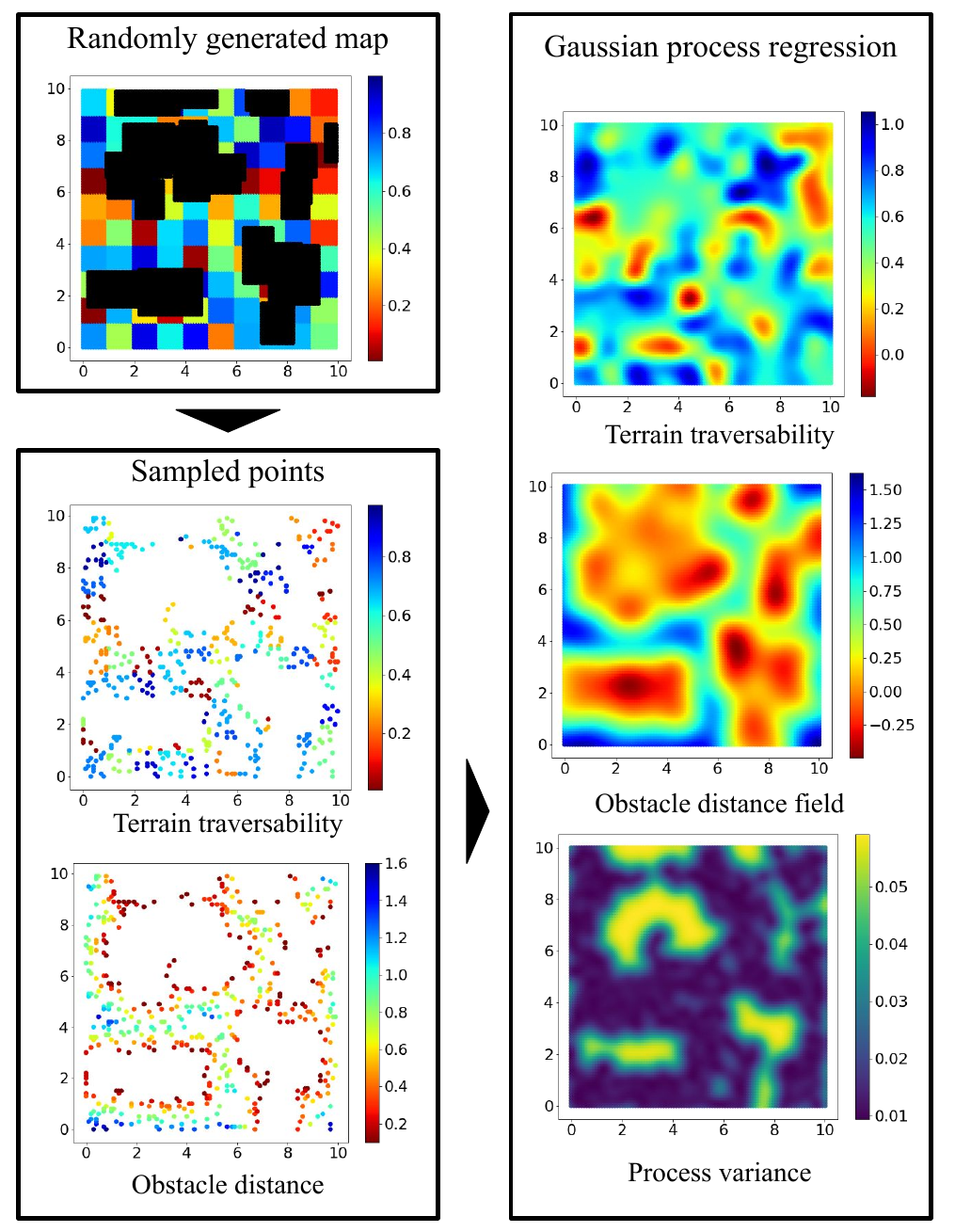}
\caption{Visualisation of a Gaussian process variational regression map over terrain traversability and obstacle distance from training data sampled in a randomly generated environment. Obstacles are displayed in black in the generated map image. \label{map_generation}}
\end{figure}

Recent work in the field of robotics employed GPR for the building of implicit representations of distance fields~\cite{online_continuous_gp} and geometric-semantic maps \cite{semantic_gp_mapping1,semantic_gp_mapping2}. The results of these developments suggest that they can be employed to provide a representation specifically adapted to robot navigation in unstructured environments. 
Embedded 3D sensors and pre-processing pipelines such as semantic segmentation and clustering can provide valuable data representing the environment properties for navigation including the terrain topography, obstacle distance fields and traversability (derived from the terrain semantic classes and geometry as in \cite{traversability_analysis}) in the form of sparse point clouds. Using these point clouds as training data, Gaussian processes can provide their regression online as demonstrated in \cite{gp_map_2017}. They also exhibit interesting properties such as continuity, differentiability and provide a process variance that can be used to model the regression uncertainty. In this study, the GPR training inputs consist of a set of 2D points $X\in\mathbb{R}^2$, each input being associated to a terrain traversability $\boldsymbol{T}\in]0,1]$ and a distance to obstacles $\boldsymbol{d}\in\mathbb{R}^+$. The average value of these outputs can be inferred for any 2D query point $q\in\mathbb{R}^2$ as follows.
\begin{align}
    \bar{T}(q) = K(q,X)^T \cdot [K(X,X)+o_n^2 \cdot I]^{-1} \cdot \boldsymbol{T} \label{posterior_mean}\\
    \bar{d}(q) = K(q,X)^T \cdot [K(X,X)+o_n^2 \cdot I]^{-1} \cdot \boldsymbol{d} \nonumber
\end{align}
Where $K(X,X)$ is the correlation matrix between input points and $K(q,X)$ the correlation vector between the query point and the inputs, both computed using a chosen correlation kernel $k(.,.)$ and $o_n$ the observation noise hyper-parameter. The process variance can be inferred at $q$ as follows.
\begin{equation}\label{posterior_variance}
    {\sigma}^2(q) = K(q,q) - K(q,X)^T \cdot [K(X,X)+o_n^2 \cdot I]^{-1} \cdot K(q,X)
\end{equation}
An example of the inputs and outputs of such a GPR is displayed in Figure \ref{map_generation}.
While the training of the GPR (i.e. the calculation of the correlation matrix $K(X,X)$ and its inversion) can be computationally heavy for large populations of training points, care must be taken into limiting their number as in \cite{gp_pseudo_inputs} or by compressing the input point clouds information using clustering for example. The calculation of the posterior distribution (\eqref{posterior_mean},\eqref{posterior_variance}) can also represent a computational bottleneck and methods of matrix decomposition or approximation can be employed to overcome this limitation as discussed further in \ref{implementation}. While the availability of such maps represents a major requirement for this study, their construction from sensor data is outside this paper scope and will not be evaluated. For further details regarding this topic, please refer to \cite{semantic_gp_mapping1,semantic_gp_mapping2,gp_map_2017}.

\section{Continuous Path Planning Using a GPR Map}\label{path-planning}
The continuous environment representation set by the GPR on the chosen variables described above allows to exploit smooth planning methods to provide safe and efficient paths for a mobile robot to follow. We thus propose to perform such path planning between a start point $s$ and a goal point $g$ inside the environment represented by the GPR, with constraints on obstacle avoidance and curvature while trying to maximize terrain traversability and minimize process variance along a path of shortest length. Gradient descent optimisation of the control points of a Bézier curve has been employed for this matter, with the additional impact study of the availability of a prior path generated by either a discrete or a sampling-based method.
\subsection{Discrete and Sampling-Based Priors}
\subsubsection{A*}
The A* discrete path-planning method \cite{a_star_the_origin} has been long used and will not be presented in details here. While carrying all the disadvantages of discrete path-planning methods such as fixed resolution and no consideration for the generated path compliance to the robot kinematics, it has the advantage of simplicity and guarantees the optimality of the generated path according to its cost function. The GPR can be inferred on a discrete grid of fixed resolution in which each node $n$ is associated to a traversability value $T(n)$ and a process variance $\sigma^2(n)$. Informative A*, as described in \cite{achat2022} and \cite{smana} can thus be employed to compute paths taking the terrain traversability and the process variance into account. Terms depending on the traversability $T(n)$ and the GPR variance $\sigma^2(n)$ of the terrain at node $n$ are thus added to the transition cost from its parent node $p$.
\begin{equation}
    c(p\rightarrow n) = |[p,n]| + f_T.(1-T(n)) + f_\sigma.\sigma^2(n)
\end{equation}
With $f_T$ being the traversability cost factor and $f_\sigma$ the variance cost factor, parameters of the method. 
Paths generated by this vastly used method can be used as baseline to be compared to the followings.
\subsubsection{T-RRT}
Rapidly exploring Random Trees (RRT)\cite{rrt_the_origin} are a vastly employed sampling-based method for path planning. It can be applied both in discrete or continuous spaces and operates a stochastic incremental search of a feasible path by sampling random points and building a tree from a given start point until reaching the goal point $g$. While it is usually faster to compute than the A* and does not requires the building of a discrete graph to be employed, the provided path has no guarantee to be optimal. The transition based RRT (T-RRT) variant, as defined in \cite{t_rrt} uses stochastic rejection of nodes based on their cost $c(n)$ to guide the search toward low-cost regions, thus being able to converge faster and provide a more efficient path. The cost of a node $n$ to be considered by the T-RRT algorithm in our context is defined similarly as for the A* method.
\begin{equation}
    c(n) = |[n,g]| + f_T.(1-T(n)) + f_\sigma.\sigma^2(n)
\end{equation}
\subsection{Bézier Curve Optimisation (BCO) by Gradient Descent}
Bézier curves are flexible parametric differentiable functions that can be used to generate smooth continuous paths for mobile robots. They are defined for a parameter ${t\in[0,1]}$ by a set $P$ of $m$ control points ${P_i=[x_i\quad y_i]^T,i\in\{0,...,m-1\}}$, providing a parametric curve in the 2D plane. At time $t$, a 2D point $(x,y)=B(t,P)$ can be inferred as follows.
\begin{equation}
    B(t,P) = \sum_{i=0}^{m-1} \frac{m!}{i!\,(m-i)!}\,t^i(1-t)^{m-i}P_i
\end{equation}
The position of these control points can be optimised to fit the loss function defined by the Gaussian process regression in a variety of manners. Since the Gaussian process provides a cheap-to-evaluate differentiable regression  with respect to its query points, gradient descent optimisation seem to be an appropriate choice for this application. 
Such an optimisation problem can be classically formulated as finding the minimum of the following loss function.
\begin{equation}
    \min_{P\in\mathbb{R}^{2m}} L(B(t,P))
\end{equation}
With $L(B(t,P))$ being the differentiable loss function of the Bézier curve in the GPR representation of the environment defined in \eqref{total_loss} as a sum of the losses associated to each characteristic of the robot navigation. 
\begin{equation}\label{total_loss}
    L(B(t,P)) = L_l + f_T.L_T + f_\sigma.L_\sigma + f_o.L_o + f_c.L_c
\end{equation}
$L_l$, $L_T$, $L_\sigma$, $L_o$ and $L_c$ correspond to the losses related respectively to the path length normalised by the distance between the start $s$ and goal $g$, the average terrain traversability and variance along the path, its collision with obstacles and curvature above a fixed threshold, defined in \eqref{loss_terms}. $f_T$, $f_\sigma$, $f_o$ and $f_c$ are the coefficients associated to each loss term, chosen in order define a hierarchy between all terms. Indeed, as gradient descent methods do not explicitly allow the introduction of constraints in the optimisation process, a penalization approach by order of importance is adopted with a significantly higher weight for obstacle avoidance, followed by path curvature.
\begin{gather}
    L_l = \int_{0}^{1} \sqrt{x'^2+y'^2}\enspace dt/\sqrt{(g-s)^2} \nonumber\\
    L_T = \int_{0}^{1} 1-\bar{T}(x,y)\enspace dt \nonumber\\
    L_\sigma = \int_{0}^{1} \sigma^2(x,y)\enspace dt \label{loss_terms}\\
    L_o = -\int_{0}^{1} \min(0,R-\bar{d}(x,y))\enspace dt \nonumber\\
    L_c = \int_{0}^{1} \min\left(0,\frac{1}{r_0}-\frac{|x'y''+y'x''|}{x'^2+y'^2}\right)\enspace dt \nonumber
\end{gather} 
$\bar{T}(x,y)$, $\sigma^2(x,y)$ and $\bar{d}(x,y)$ are respectively the mean terrain traversability, mean distance to obstacles and process variance inferred by the GPR at coordinates $(x(t),y(t))$. $R$~is the robot safety radius and $r_0$ the minimum acceptable curvature of the path. 

The number and initial coordinates of the Bézier curve control points need to be initialised. A straightforward way of setting these initial values is to uniformly place the control points along a straight line between the start and goal of the planner at a given spatial resolution $cpr$. However, as such optimisation methods tend to fall in local minima, the use of a prior over the control points could be helpful to find a feasible and low-cost path, if such a path exists. Moreover, the complexity of the optimisation process is highly dependent on the number of control points. While too little control points could lead to the optimisation failure in finding a feasible path, too many points lead to greater computational load. We thus propose to exploit the path generated by the A* and the T-RRT algorithm to provide the number of control points and their initial positions. The usage of the A* prior, T-RRT prior and priorless initialisation are compared in Section \ref{results}.

\section{Numerical Experiments}\label{evaluation}
The paths planned by A*, T-RRT, BCO with an A* prior (BCO-A*), a T-RRT prior (BCO-RRT) and without prior (BCO-$\O$) have been evaluated and compared in terms of compliance to the defined constraints and computation time using a set of procedurally generated 2D GPR environment representations. 
\subsection{Implementation}\label{implementation}
The Gaussian process regression is computed on GPU using the GPyTorch library \cite{gpytorch}. The Radial Basis Function~(RBF) has been chosen as the GPR correlation kernel, its hyper-parameters along with the observation noise $o_n$ have been optimised by minimising the GPR marginal log-likelihood with respect to its training points. The LOVE method \cite{gp-love} integrated in GPyTorch allows accelerated and precise estimation of the Gaussian process posterior variance. The latter is mostly helpful when inferring a GPR on large batches of query points at once and thus particularly efficient for building the A* discrete graph or computing the Bézier curve losses. The GPR closed-form expression allows to compute the optimisation loss function gradient with respect to the control points coordinates. The PyTorch Autograd tool is used for this matter and handles the back-propagation of the gradient through all successive operations. Gradient descent is performed using the Adam stochastic-gradient optimiser. Optimisation is performed for a given maximum of 500 iterations and an early stopping condition has been set over loss convergence (i.e. when the difference between successive loss values is below a given threshold). The integrals required for the calculation of the optimisation loss have been numerically computed using uniform sampling along the Bézier curves. The number of points sampled along the path is updated at every optimisation step according to the current path length to ensure sampling at a fixed spatial resolution, regardless of the control points coordinates. For consistency, parameters shared between the A*, T-RRT and BCO have been set to the same values. Table \ref{parameters} summarizes the roles and chosen values of all method parameters.
\begin{table}[]
\vspace{6pt}
\centering
\begin{tabular}{|l|l|l|l|}
\hline
Method & Parameter & Value  & Description                                                                                               \\ \hline
Shared   & $R$        & 0.1 m & \begin{tabular}[c]{@{}l@{}}Safety radius: threshold obstacle\\ distance value for coordinates to \\ be considered as occupied\end{tabular} \\ \cline{2-4} 
         & $f_T$      & 10    & \begin{tabular}[c]{@{}l@{}}Traversability term factor in \\ optimisation loss and node costs\\ computation\end{tabular}              \\ \cline{2-4} 
         & $f_\sigma$ & 200   & \begin{tabular}[c]{@{}l@{}}Process variance term factor in\\ optimisation loss and node costs\\ computation\end{tabular}             \\ \hline
A*     & res       & 0.1 m  & \begin{tabular}[c]{@{}l@{}}Resolution of the discrete A* \\ graph\end{tabular}                            \\ \hline
T-RRT  & step      & 0.5 m  & \begin{tabular}[c]{@{}l@{}}Maximum distance between two\\ nodes of the tree\end{tabular}                  \\ \hline
BCO & res        & 0.1 m & \begin{tabular}[c]{@{}l@{}}Spatial resolution of the sampling\\ along the path to approximate its\\ integral\end{tabular}            \\ \cline{2-4} 
       & $r_0$ & 0.25 m & \begin{tabular}[c]{@{}l@{}}Minimum acceptable radius of \\ the path curvature\end{tabular}                \\ \cline{2-4} 
       & lr        & 0.05   & \begin{tabular}[c]{@{}l@{}}ADAM optimiser initial learning\\ rate\end{tabular}                            \\ \cline{2-4} 
       & $f_o$     & 1000   & \begin{tabular}[c]{@{}l@{}}Obstacle distance term factor \\ in optimisation loss computation\end{tabular} \\ \cline{2-4} 
       & $f_c$     & 100    & \begin{tabular}[c]{@{}l@{}}Path curvature term factor in\\ optimisation loss computation\end{tabular} 
       \\ \cline{2-4} 
       & $cpr$   & 0.5 m  & \begin{tabular}[c]{@{}l@{}}Spatial resolution of the control\\ points placement for priorless\\ initialisation\end{tabular} \\ \hline
\end{tabular}
\caption{Summary of the different planning methods parameters, values and description for this study.}
\label{parameters}
\vspace{-0.5cm}
\end{table}

\subsection{Procedural Generation of GPR Maps}
A set of 50 synthetic environments has been generated to evaluate and compare the path-planning methods described in Section \ref{path-planning}. They consist of 100~m² square maps composed of 1~m² ground tiles assigned with a traversability value randomly selected in $]0,1]$. They are incrementally populated with rectangular obstacles of random sizes and positions until an occupancy threshold is reached. The latter is chosen randomly in $[0.25,0.5]$ for each map. A set of 500 training points is randomly sampled following a uniform distribution in the free space of each map. The distance to the closest obstacle is computed for each point and these distances along with their traversability and 2D coordinates are given to a GPR as training data, thus generating the environment representation in which path planning can be performed. Figure \ref{map_generation} illustrates the described map generation process.

\subsection{Evaluation \& Results}\label{results}
\begin{figure}[]
\vspace{4pt}
\centerline{}
\includegraphics[width=0.49\textwidth]{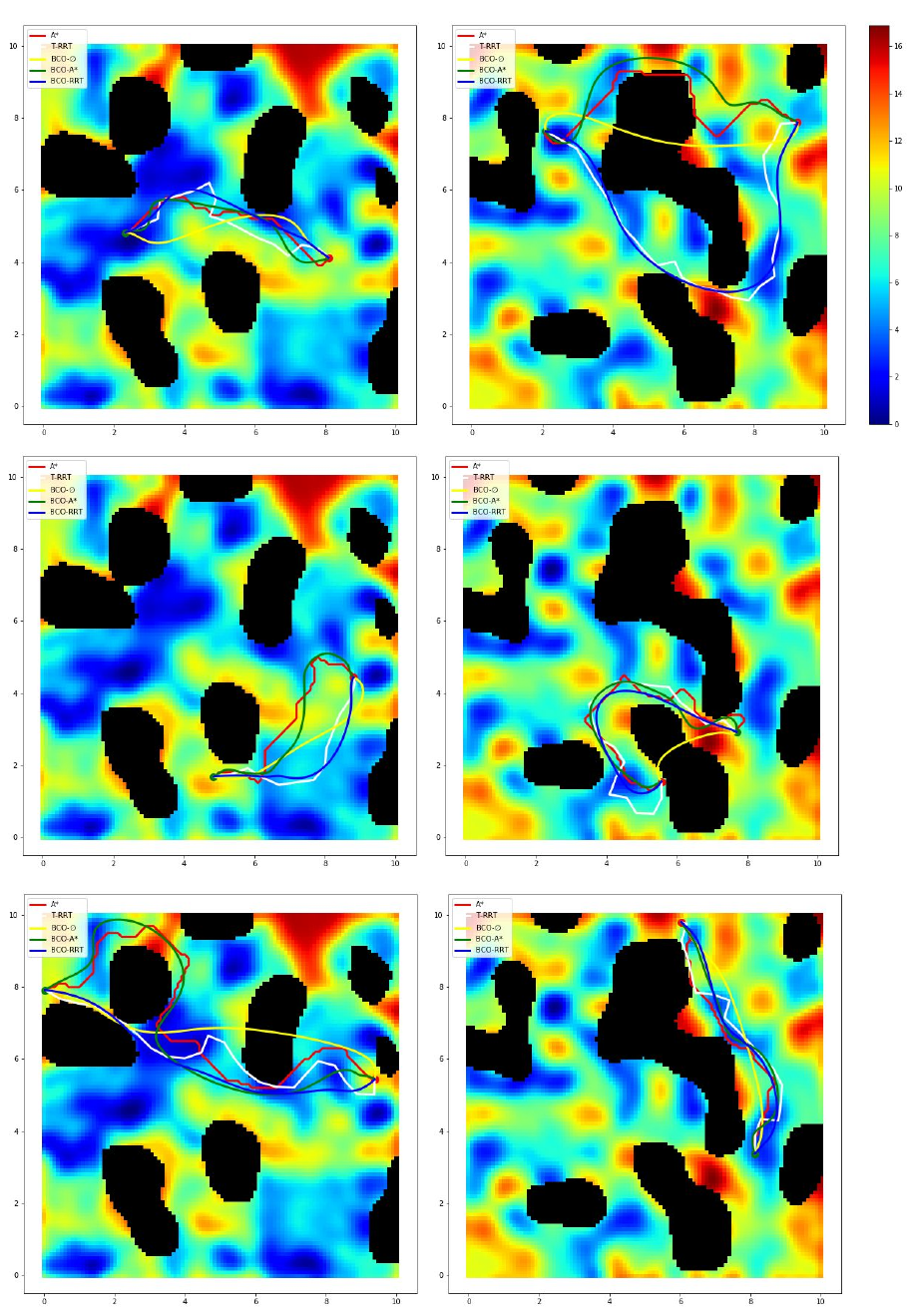}
\caption{Examples of paths computed by each evaluated method between random pairs of points in a procedurally generated GPR map. Colors represent the cumulative map related costs. Obstacles (i.e. coordinates at which $\bar{d}(x,y)\leq R)$ are displayed in black. \label{path-exemples}}
\end{figure}

\begin{figure}[]
\vspace{4pt}
\centerline{}
\includegraphics[width=0.49\textwidth]{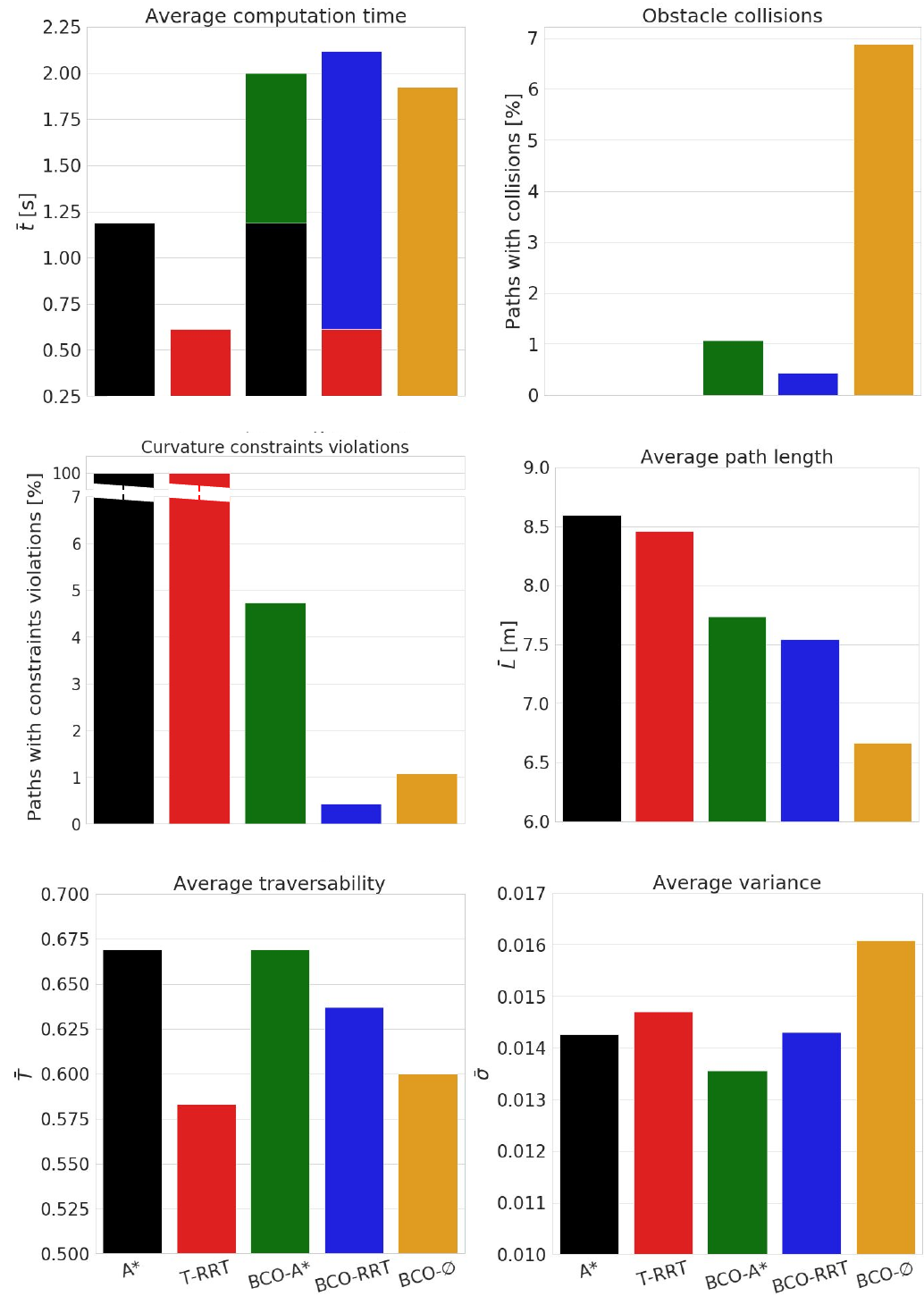}
\caption{Results of the comparison of the paths planned respectively by the A*, T-RRT, BCO-A*, BCO-RRT and BCO-$\O$. The A* and T-RRT prior computation times are respectively added to the BCO-A* and BCO-RRT optimisation times.
\label{evaluation_results}}
\end{figure}

A set of 100 pairs of start and goal points with an existing collision-free A* path between them has been selected for each of the 50 generated maps. A randomised ablation of the GPR training points has been applied to generate areas of little confidence and help the evaluation of the map variance influence over optimisation. Paths between all 5000 pairs of start-goal couples have been computed with A*, T-RRT, BCO-$\O$, BCO-A* and BCO-RRT. This evaluation was performed on an Intel Xeon(R) W-2123 8 core 3.60GHz CPU with a NVIDIA GeForce GTX 1080 GPU. Figure \ref{path-exemples} displays examples of paths generated by each method in a given GPR map. 
These paths have been evaluated and compared in terms of average computation time, length, traversability and variance, number of paths presenting at least one curvature constraint violation and number of paths colliding at least once with an obstacle. Figure \ref{evaluation_results} presents the results of this evaluation averaged over every computed path for each method. 

\begin{figure}[]
\vspace{4pt}
\centerline{}
\includegraphics[width=0.49\textwidth]{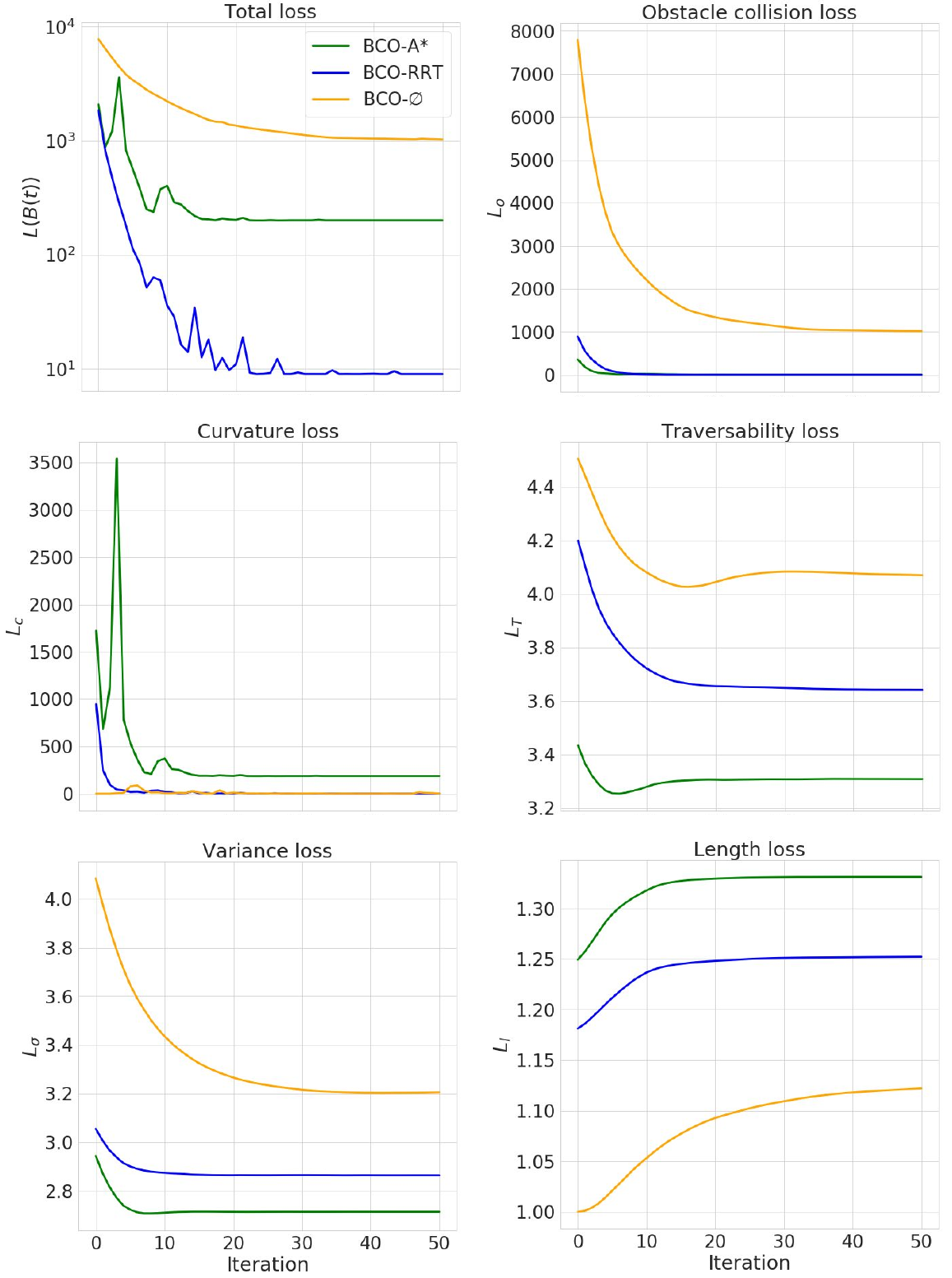}
\caption{Average over all generated paths of the total loss $L(B(t,P))$ and the different loss terms $L_o$, $L_c$, $L_T$, $L_\sigma$ and $L_l$ after each optimisation iteration for all experiments. Only 50 out of the maximum 500 iterations are displayed since all loss terms seem to have converged at this point.\label{evaluation_losses}}
\end{figure}

Both A* and T-RRT methods cannot take curvature constraints into account, all computed paths being a collection of segments, therefore they all naturally break the curvature constraint, as opposed to the BCO method using any kind of prior. The comparison between the number of paths colliding with obstacles for BCO highlights the interest of using a feasible path as a prior. As illustrated in Figure \ref{path-exemples}, many paths computed without prior end up colliding with obstacles due to the optimiser falling in local minima. All three metrics on the paths lengths, average traversabilities and variances are improved by the BCO process, with the optimised paths displaying higher traversabilities and lower variances and lengths than their respective prior. As for the comparison of the results of the optimisation process using the A* or the T-RRT prior, both display their respective advantages and drawbacks. When using the A* path as prior, the BCO seems to be slightly more prone to violating the curvature and collision requirements. 
While the calculation of the A* prior is more computationally heavy than the T-RRT, the following optimisation process is shorter in average and provides more efficient paths in terms of traversability and variance. 
All loss terms have been recorded at each iteration of the optimiser for every kind of initialisation. Figure \ref{evaluation_losses} displays their average values over every computed path. These results show that the optimisation process is successful and that its convergence is reached before the fixed limit of 500 iterations. The hierarchy between the average total losses of the three evaluated methods can be explained by the fact that BCO-$\O$ often fails to respect the collision constraints and by BCO-A* slight tendency to break the curvature constraints.

\section{Conclusion \& Perspectives}
The gradient descent optimisation of a Bézier curve control points presented in this paper has been demonstrated to be applicable for efficient smooth path planning using a Gaussian process regression of the terrain traversability and obstacle distance fields as the environment representation. The interest of using a coarse feasible path as prior for the optimisation has been demonstrated, both in terms of paths quality and computation speed. The whole process requires reasonable computation power and should thus be applicable for online embedded path planning. In order to improve the performances and robustness of the planning, projected gradient descent \cite{projected_gradient} could be used to transform the described methodology into a constrained optimisation problem. Methods of global optimisation including evolutionary strategies (such as CMA-ES \cite{cma-es}) could allow to find a globally optimal path and bypass the need for a prior, but at a higher computational cost. The number of control points of the Bézier curve could also be jointly optimised, although this represents a more challenging problem. Nonetheless, these promising results pave the way to the integration of such planning methods in a fully autonomous robotic perception and navigation pipeline that could be deployed on a ground robot to perform navigation tasks in real-world challenging environments.

\bibliographystyle{ieeetr} 
\bibliography{bibliographie}

\end{document}